\documentclass[conference]{IEEEtran}
\IEEEoverridecommandlockouts
\usepackage{cite}
\usepackage{amsmath,amssymb,amsfonts}
\usepackage{algorithmic}
\usepackage{graphicx}
\usepackage{textcomp}
\usepackage{xcolor}
\usepackage{float}
\usepackage{array}
\usepackage{multicol}
\usepackage{makecell}
\usepackage{siunitx}
\usepackage{multirow}
\usepackage{subfig}
\usepackage{comment}
\usepackage{gensymb}
\usepackage{pdfpages}
\usepackage{hyperref}
\usepackage{mathtools}
\usepackage[ruled,vlined]{algorithm2e}

\def\BibTeX{{\rm B\kern-.05em{\sc i\kern-.025em b}\kern-.08em
    T\kern-.1667em\lower.7ex\hbox{E}\kern-.125emX}}
\begin{document}
\newcommand{\myfigref}[2]{~\ref{#1}.\subref{#2}}

\title{Symbolic regression for scientific discovery: an application to wind speed forecasting\\
}

\author{\IEEEauthorblockN{Ismail Alaoui Abdellaoui}
\IEEEauthorblockA{\textit{Department of Data Science and Knowledge Engineering} \\
\textit{Maastricht University}\\
Maastricht, The Netherlands \\
i.alaouiabdellaoui@student.maastrichtuniversity.nl}
\and
\IEEEauthorblockN{Siamak Mehrkanoon*\thanks{*Corresponding author.}}
\IEEEauthorblockA{\textit{Department of Data Science and Knowledge Engineering} \\
\textit{Maastricht University}\\
Maastricht, The Netherlands \\
siamak.mehrkanoon@maastrichtuniversity.nl}
}

\maketitle

\begin{abstract}
Symbolic regression corresponds to an ensemble of techniques that allow to uncover an analytical equation from data. Through a closed form formula, 
these techniques provide great advantages such as potential scientific discovery of new laws, as well as explainability, feature engineering as well as fast inference. Similarly, deep learning based techniques has shown an extraordinary ability of modeling complex patterns. The present paper aims at applying a recent end-to-end symbolic regression technique, i.e. the equation learner (EQL), to get an analytical equation for wind speed forecasting. We show that it is possible to derive an analytical equation that can achieve reasonable accuracy for short term horizons predictions only using few number of features. 

\end{abstract}


\section{Introduction}
A great amount of natural phenomena are explained and computed through a short mathematical expression. For instance, the phenomenon of gravity is explained through Newton's laws of motion. Similarly, the mass-energy equivalence is described through Einstein's famous formula. Those equations generally contain a low number of terms and constants, allowing us to understand the relationship between them. 
At the same time, machine learning data-driven based models have made a rapid progress in several fields, including computer vision, computational linguistics, and dynamical systems \cite{mehrkanoon2012approximate,voulodimos2018deep,manning2015computational, mehrkanoon2015learning, liu2021hybrid, mehrkanoon2012ls, wang2021ultra, mehrkanoon2016estimating, da2021novel,mehrkanoon2014parameter, mehrkanoon2019deep,mehrkanoonbroadUnet, mehrkanoon2014large,moreno2021hybrid,mehrkanoon2019cross,li2021using,Tomasz}.

Symbolic regression aims at searching the space of mathematical expressions that best fit a given dataset. The main motivation behind this approach is to get an interpretable model that offers an alternative to the black-box models such as neural networks. Thanks to the interpretable nature of symbolic regression based models, they have been widely used in industrial empirical modeling \cite{kotanchek2008trustable,yang2015modeling,sun2019data,vazquez2020combination}. Due to the high dimensional space of mathematical expressions that can describe a specific dataset, symbolic regression is a complex combinatorial problem \cite{benabbou2020interactive,basmassi2020novel}. Therefore, the traditional symbolic regression approaches have been mostly based on genetic algorithms. However, it has been recently shown that genetic programming applied to symbolic regression is unable to scale to large systems, and might overfit the data \cite{kim2020integration}.

On the other hand, advanced machine learning techniques such as deep learning has shown their great potential to address large problems. In particular, deep learning based models applied to weather data has recently shown a lot of success due to the rapid advancement of data-driven modeling \cite{mehrkanoon2019deep2,trebing2020smaat,trebing2020wind,barajas2019performance, fernandez2020deep,zaytar2016sequence}. Some other work also investigated explainability techniques aimed at weather forecasting using deep learning. These techniques include occlusion analysis and score maximization \cite{abdellaoui2020deep,mahendran2016visualizing,molnar2020interpretable}. The present paper aims at extending the work in \cite{kim2020integration} and applying it to real weather data that includes a high number of input features. Furthermore, we study to which extent can we obtain an analytical equation with a low number of features and terms, while maintaining a reasonable accuracy. More specifically, here we focus on the wind speed forecasting task for three Danish cities. The same dataset as in \cite{mehrkanoon2019deep2} is used and the obtained results of the proposed model are compared with those of \cite{mehrkanoon2019deep2}. In particular we compare our results with the best performing model in \cite{mehrkanoon2019deep2}, which is based on a 3d convolutional neural network.

This paper is organized as follows. A brief overview of the existing symbolic regression methodologies is given in Section \ref{sec:related_work}. The description of the EQL network introduced in \cite{kim2020integration} is presented in \ref{sec:preliminaries}. Furthermore, the dataset used is introduced in Section \ref{sec:data_desc}. The experimental results are reported in section \ref{sec:results}. Finally, a discussion followed by the conclusion are drawn in sections \ref{sec:discussion} and \ref{sec:conclusion}, respectively.

\section{Related Work}\label{sec:related_work}
There are many approaches proposed in the literature for symbolic regression problem. From a traditional standpoint, genetic programming has always been used for symbolic regression due to the large search space for possible analytical equations. 
The authors in \cite{smits2005pareto}, introduced ParetoGP methodology that exploit the Pareto front to address the trade-off between the accuracy of the prediction and the complexity of the equation (e.g. the total number of terms).

Indeed, models that result from unconstrained genetic programming can be accurate, however resulting in a tree structure that is too deep or that incorporates a large number of nodes. This multi-objective optimization approach proved to yield superior results compared to previous approaches. Additionally, the work presented in \cite{stijven2016prime, masmoudiartificial,gasper2021challenging} show some real world applications of symbolic regression using genetic algorithms, including business forecasting (e.g. labor cost and product demand) for decision making, as well as complex systems analysis (e.g. spread of infectious diseases). Similarly to this work, the authors in \cite{sun2019data} and \cite{wang2019symbolic} make an extensive list of industrial applications by means of symbolic regression with genetic programming. These applications include the Hamiltonian and Lagrangians of oscillating systems, the governing equations of gas and wind turbines, systems for pipe failures among others. The work in \cite{wang2019symbolic} states the opportunities of symbolic regression in material science by predicting the changes of materials properties and performance in response to perturbations. 

While genetic algorithms have often been used to find mathematical expressions, it has been shown that this methodology is prone to overfitting, and is unable to scale to large systems \cite{kim2020integration}. Therefore, more recent approaches use a combination of neural networks and genetic programming. In the work proposed by Cranmer et. al, Graph Networks \cite{battaglia2018relational} are used to uncover known physics equations and to discover an unknown cosmology law \cite{cranmer2020discovering}. This type of neural network has been chosen because of its strong inductive bias.

In particular, data from moving particles the Graph Network was used to predict the instantaneous acceleration of each particle in the system. To encourage a compact mathematical expression, a sparsity constraint was used during the training to reduce the dimensionality of each node in the graph. In the second phase of the training, the input data as well as its corresponding prediction by the network is fed to the Eureqa software \cite{dubvcakova2011eureqa} to find the final analytical expression.

The most recent advances in symbolic regression leverage the predicting power of neural networks. The work presented in \cite{champion2019data} uncovers a Lorenz system by means of incorporating an autoencoder and using some components of the  Sparse Identification of Nonlinear Dynamics (SINDy) approach \cite{brunton2016discovering}. This methodology proved to be successful, since only an affine transformation of the result was necessary to uncover the original Lorenz system. However, it should be noted that one of the  requirements of this methodology is an a-priori knowledge about the dimensionality of the autoencoder's bottleneck. Another recent work by Kim et al. in \cite{kim2020integration}, made use of a simple feed forward network, the equation learner (EQL) to extract a mathematical expression given an arbitrary dataset. Some particularities of the EQL approach is that each layer in the network is made of multiple types of activation functions and the training includes a specific regularization procedure. One particular sought after advantage of EQL is the compactness of the final equation as well as the automatic identification of the most important input features. However, this study was only applied to a relatively low number of inputs.

\section{Preliminaries}\label{sec:preliminaries}
\subsection{EQL approach} \label{ssec:eql_layer}
\begin{figure}[H]
\centering
\subfloat[]{{\includegraphics[width=\columnwidth]{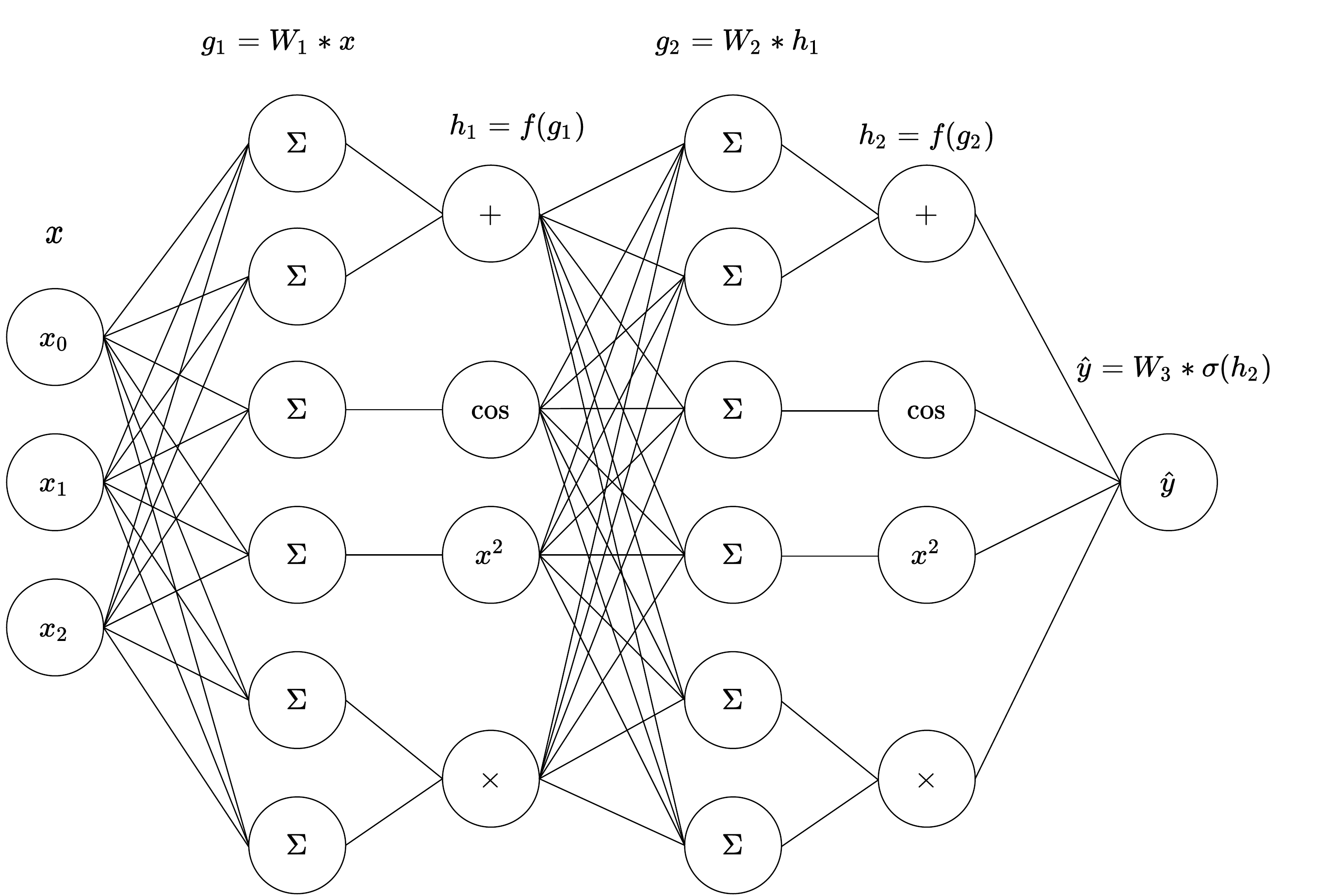}}}
\caption{Schema of the EQL network introduced in \cite{kim2020integration}.}
\label{fig:eqlnet}
\end{figure}

The EQL layer architecture is mainly based on a dense layer, yet incorporates some elements designed for symbolic regression \cite{kim2020integration}. Let us consider a 2 layer EQL network. Starting from the input vector $x$, we get an intermediate vector $g_1$ which is the projection of $x$ on the weight matrix $W_1$. The next step consists of using a list of activation functions $f$ that operate on $g_1$ to obtain $h_1$. These activation functions act as primitive functions for the final analytical expression. The way these functions operate on $g_1$ will depend whether each function is a unary operator (e.g. $\cos(\cdot)$, $\sin(\cdot)$, $(\cdot)^2$, etc.) or a binary operator (e.g. `$+$', `$-$', `$\times$', etc.). Moreover, it is advised in \cite{kim2020integration} to repeat the same function multiple times within a layer to escape local minima during training. The computations of $g_1$ and $h_1$ form the 1\textsuperscript{st} layer of this network. We then get $g_2$ by multiplying $W_2$ and $h_1$. The vector $h_2$ is computed by applying the same list of activation functions $f$. Finally, we get the prediction $\hat{y}$ by multiplying $W_3$ and the result of the sigmoid function on $h_2$. A schema of the EQL network is depicted in Fig. \ref{fig:eqlnet}, with the operators `$+$', $\cos(\cdot)$, $(\cdot)^2$, and `$\times$' as the activation functions $f$. The training procedures consists of 2 phases. The first phase uses a particular type of regularization that will be explained in the next section. During the second phase, we first perform a thresholding that will set weights below a certain value at 0. Then the zero-valued weights are frozen while another training occurs that does not make use of regularization. 

\subsection{Smoothed $L_{0.5}$ regularization}
\begin{figure}[H]
\centering
\includegraphics[width=\columnwidth]{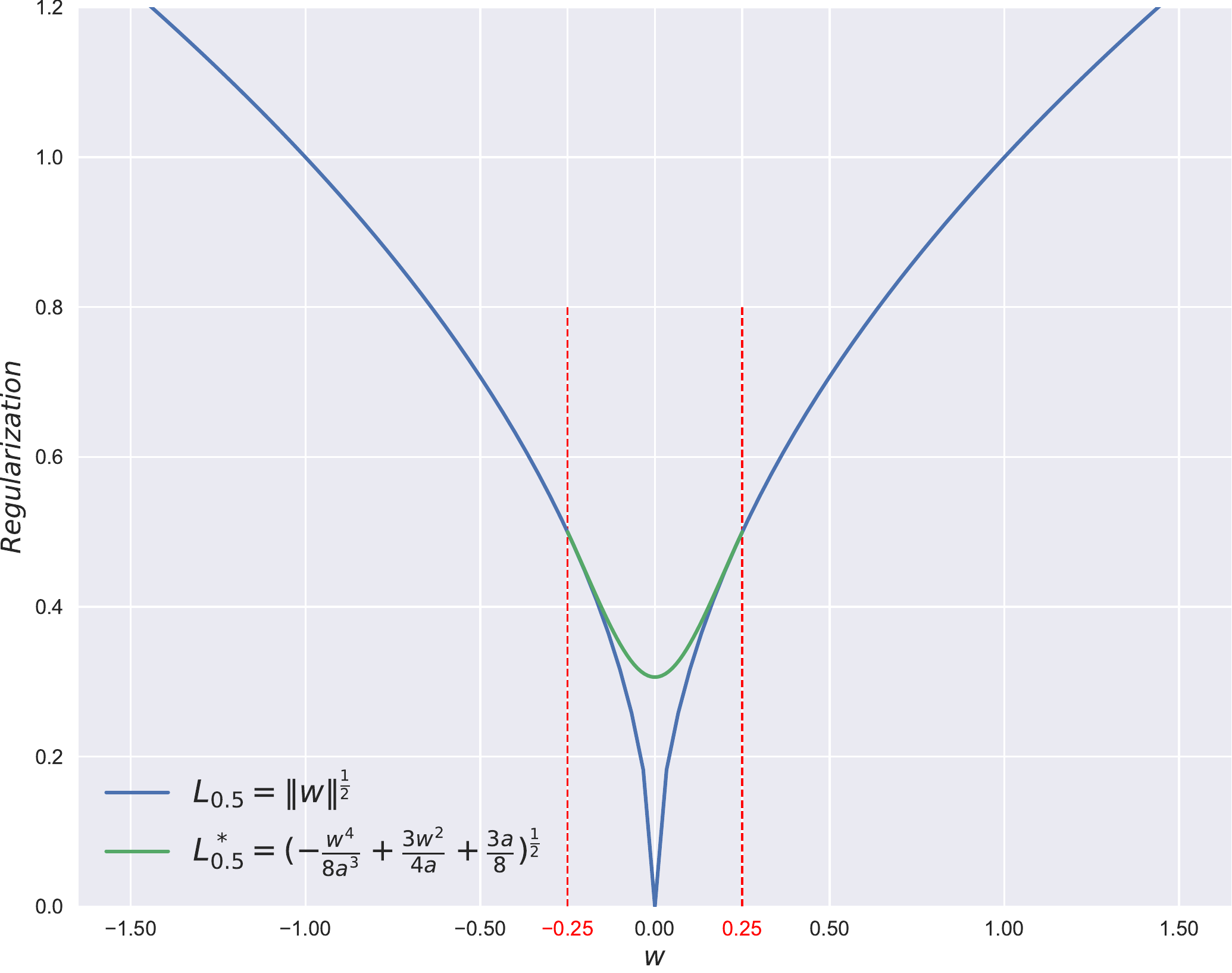}
\caption{Comparison of the $L_{0.5}$ and $L_{0.5}^*$ regularizations, for illustration purposes, we set $a$ equal to 0.25.}
\label{fig:reg_compare}
\end{figure}

To encourage sparsity and to avoid overfitting, regularization was another key element of the EQL approach \cite{kim2020integration}.
It was shown that the conventional regularization penalties were not effective for the EQL network and that using a $L_{0.5}$ regularization was leading to simpler expressions than using $L_0$ or $L_1$ regularization. However in practice, the authors in \cite{kim2020integration} used $L_{0.5}^*$, a smoothed version of the $L_{0.5}$, because it empirically led to a more stable gradient descent. More specifically, the weights are computed as follows:
\[
    L_{0.5}^*(w)= 
\begin{cases}
    \Vert{w}\Vert^{\frac{1}{2}}& \text{if } \Vert{w}\Vert\geq a\\
    (-\frac{w^4}{8a^3}+\frac{3w^2}{4a}+\frac{3a}{8})^{\frac{1}{2}}& \text{if } \Vert{w}\Vert < a
\end{cases}
\]
We should note that the parameter $a$ is a positive real number, and serves as a threshold point between the regular and the smoothed regularization term. Fig. \ref{fig:reg_compare} illustrates the comparison between the two regularizations.

\subsection{Baseline model}\label{ssec:basline_model}
The 3d convolutional neural network model introduced in \cite{mehrkanoon2019deep2} will serve as baseline model. The input of this model is a tensor $\mathcal{T} \in \mathbb{R}^{C \times L \times F}$, where $C$ is the number of cities, $L$ is the number of lags used, and $F$ is the number of weather features. This tensor is then fed to a 3D convolutional layer, with a kernel size of (2$\times$2$\times$2). Next the $ReLu$ activation function is applied and the resulting volume is then flattened to be fed to a dense layer. Similarly to the 3D-CNN layer, the dense layer also uses a $ReLu$ activation function. The last layer used for the prediction uses a linear activation function.

\section{Data Description}\label{sec:data_desc}
Here we use the same wind speed dataset as used in \cite{mehrkanoon2019deep2}, which is publicly available \footnote{\url{https://sites.google.com/view/siamak-mehrkanoon/code-data?authuser=0}}. The dataset used originates from the National Climatic Data Center (NCDC) and concerns 5 Danish cities and 4 weather features spanning from 2000 to 2010. The time resolution of this dataset is hourly, and the weather features include the temperature, the pressure, the wind speed, and the wind direction. The 5 Danish cities are Aalborg, Aarhus, Esbjerg, Odense, and Roskilde. Each time step $t$ is defined by a matrix $M_t\in \mathbb{R}^{F \times C}$, where $F$ is the number of features and $C$ is the number of cities. Therefore the whole dataset is a tensor $\mathcal{D} \in \mathbb{R}^{L \times F \times C}$, where $L$ is the total number of hours used.

\begin{figure}[H]
\centering
\includegraphics[width=\columnwidth]{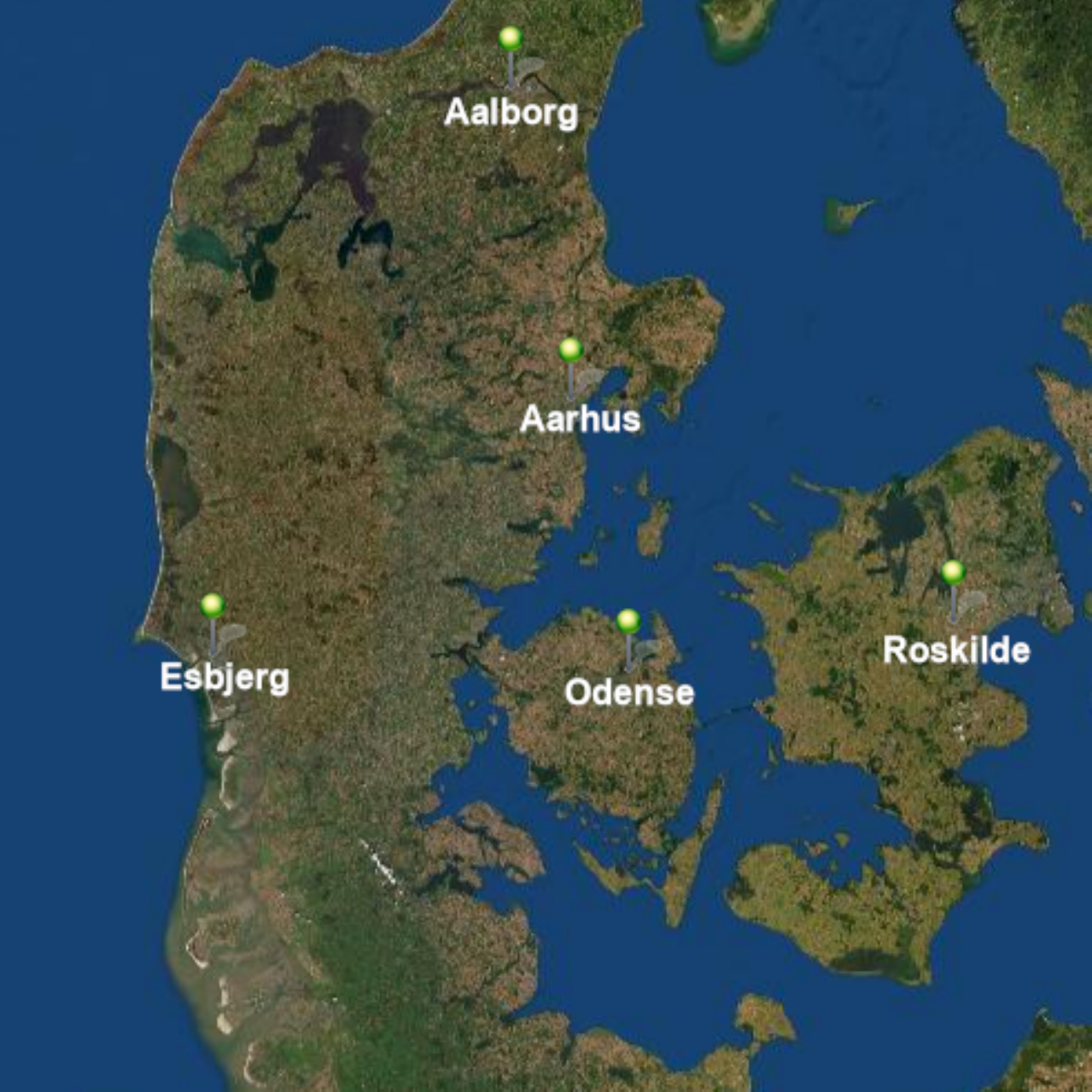}
\caption{Map of the 5 Danish cities used in the dataset.}
\label{fig:map_data}
\end{figure}

\section{Experimental Results}\label{sec:results}
\subsection{Data Preprocessing}\label{sec:Preprocessing}
The weather data is first scaled by means of equation \ref{eq:minmax}. In this way, for each feature and city, we take the values corresponding to every date and scale them down between 0 and 1.
\begin{equation}\label{eq:minmax}
    x_{scaled} = \frac{x - min(c_{ij})}{max(c_{ij}) - min(c_{ij})},
    i \in [1,F], j \in [1,C],
\end{equation}
where $c_{ij} \in \mathbb{R}^{L}$ refers to a column vector and $i$ and $j$ are the i\textsuperscript{th} feature of the j\textsuperscript{th} city.

\subsection{Experimental setup}

For all the experiments conducted, we focus on the following three target cities: Esbjerg, Odense, and  Roskilde. Moreover, since the EQL network is not a multi-output network, we train the network for each city separately. The target feature selected is the wind speed. Following the lines of \cite{mehrkanoon2019deep2},
we perform 6 hours ahead preddiction for each city and the lag value is set to 4.  Therefore, each data input sample at time $t$ is a vector $x_t\in \mathbb{R}^{F\times C \times L}$ of dimension 80, since there are 5 input cities, 4 weather features, and 4 lags.

In all the experiments, we use $90\%$ of the data for training, while the remaining $10\%$ is used for validation. Rmsprop algorithm \cite{tieleman2012lecture} is used to optimize the loss $L_{EQL}$ shown in equation \ref{eq:loss}. The regularization term is only used in the first phase of the training, i.e. $\lambda$  is set to zero for the second phase. The first phase of the training uses a learning rate of $1e^{-4}$ and the second one a learning rate of $1e^{-5}$. Both phases are trained for 100 epochs and use a batch size of $200$. All the experiments use a 2 layer network, which contains a total number of 1744 learnable parameters. The values of $\lambda$, $a$, and the threshold are hyperparameters specific to each target city. In order to find the most suitable hyperparameters for each target city, a grid search over different $\lambda$ and 
$a$ values is performed.

\begin{figure}[H]
\centering
\includegraphics[width=\columnwidth]{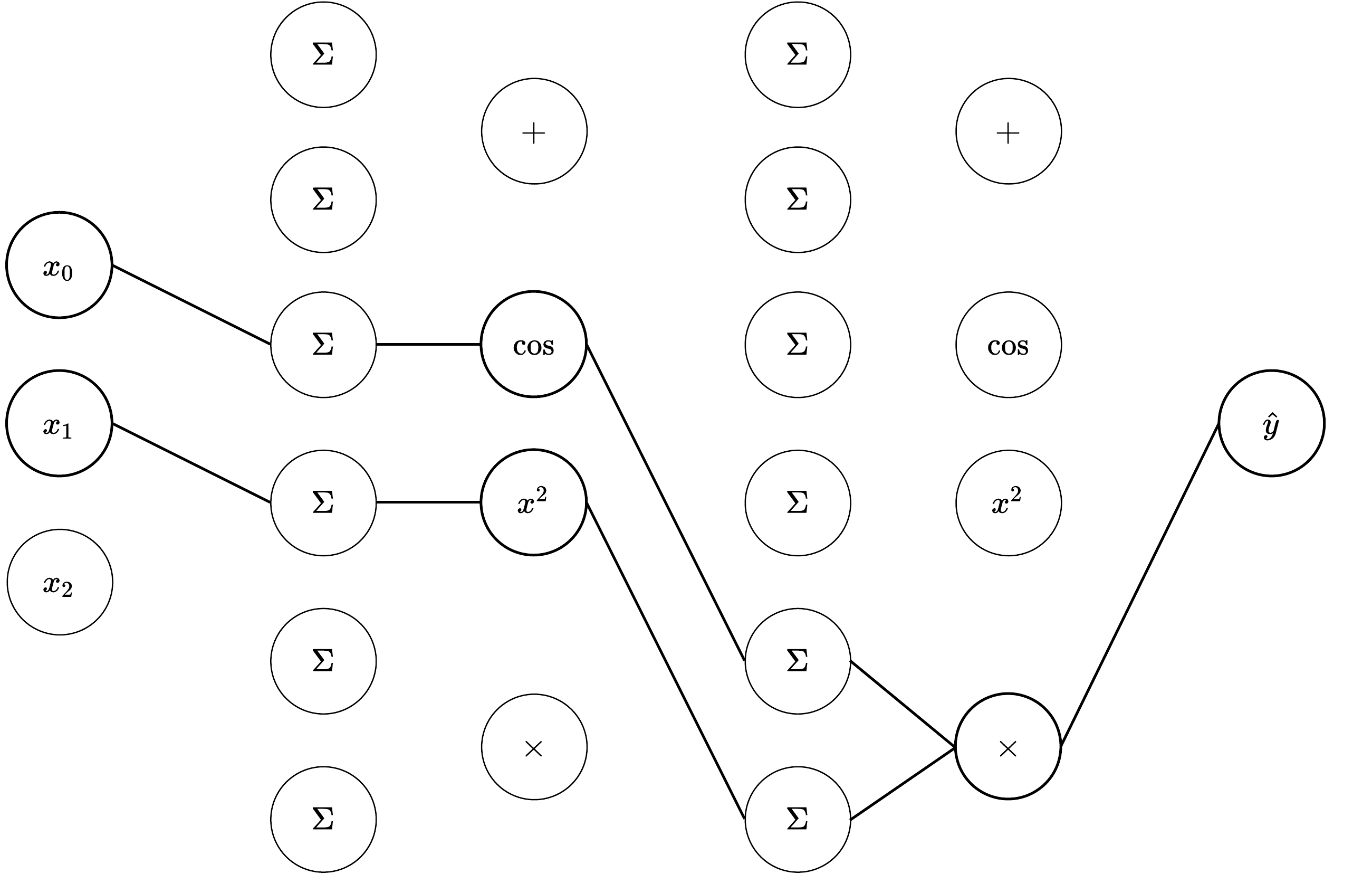}
\caption{EQL network after thresholding and regularization. For this particular example, the prediction $\hat{y}$ is described by the equation $x_1^2\cos{(x_0)}$.}
\label{fig:eqlnet_after_training}
\end{figure}

Table \ref{tab:hyperparameters} tabulates the best obtained hyperparameters. In this table, the column $S$ shows the sparsity percentage of the whole network, meaning the number of zero-valued weights divided by the total number of weights. Different levels of sparsity correspond to different thresholding values. The columns $S_1$, $S_2$, and $S_3$ in Table \ref{tab:hyperparameters}, show the sparsity of the weight matrices $W_1$, $W_2$, and $W_3$ described in section \ref{ssec:eql_layer}. Here, the list of activation functions used is the same as in \cite{kim2020integration}, except the exponential function since it was empirically degrading the performance. Fig. \ref{fig:eqlnet_after_training} depicts an example of EQL network after the training of the second phase, yielding the equation of $x_1^2\cos{(x_0)}$. 
\begin{equation}\label{eq:loss}
    L_{EQL} = \frac{1}{n}\sum_{i=1}^{n} (y_i-\hat{y}_i)^2 + \lambda L_{0.5}^*(w,a)
\end{equation}

\begin{table}[H]
    \centering
    \caption{Hyperparameters used during the training for each target city.}
    \label{tab:hyperparameters}
    \resizebox{\columnwidth}{!}{%
    \begin{tabular}{c c c c c c c c}\Xhline{3\arrayrulewidth}
    \multirow{2}{*}{City}&
    \multirow{2}{*}{\textbf{$\lambda$}}&\multirow{2}{*}{\textbf{$a$}}&
    \multirow{2}{*}{Threshold}&\multirow{2}{*}{\textbf{$S$}}&
    \multirow{2}{*}{\textbf{$S_1$}}&\multirow{2}{*}{\textbf{$S_2$}}&
    \multirow{2}{*}{\textbf{$S_3$}}\\
    & & & & & & &\\\Xhline{3\arrayrulewidth}
       Esbjerg & 5 & 5e-3 & 8.0e-3 & 98 & 99.86 & 91.67 & 43.75\\
       Odense & 5 & 5e-4 & 7.5e-3 & 98 & 99.58 & 92.36 & 50.00\\
       Roskilde & 3 & 5e-3 & 7.5e-3 & 98 & 99.72 &94.4 & 56.25\\
        \hline
        \Xhline{3\arrayrulewidth}
    \end{tabular}
    }
\end{table}
\subsection{Results} \label{ssec:results}
\begin{figure}[!htbp]
\centering
\subfloat[]{{\includegraphics[scale=0.4]{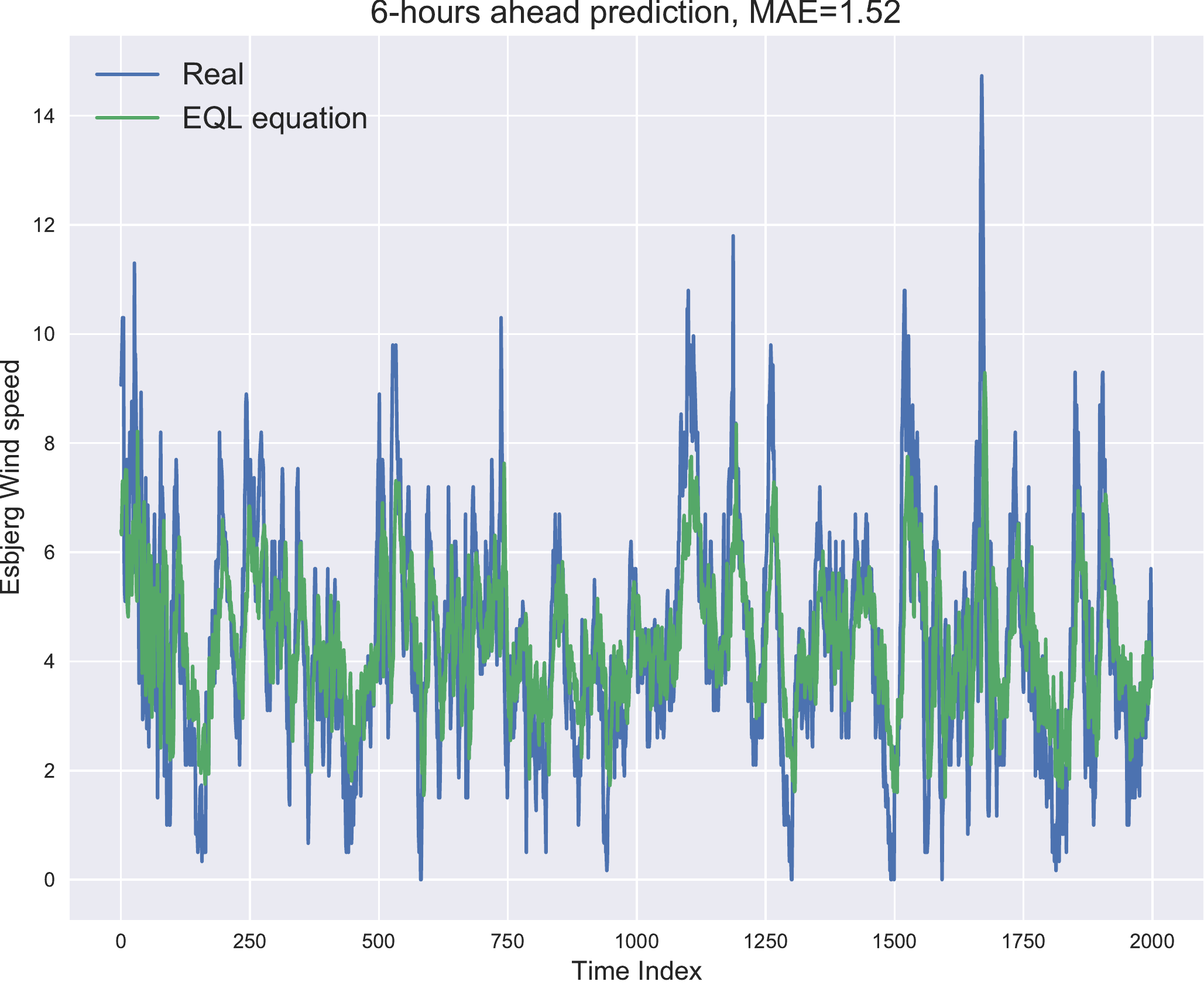}}}
\\\vskip 0.5pt plus 0.25fil
\subfloat[]{{\includegraphics[scale=0.4]{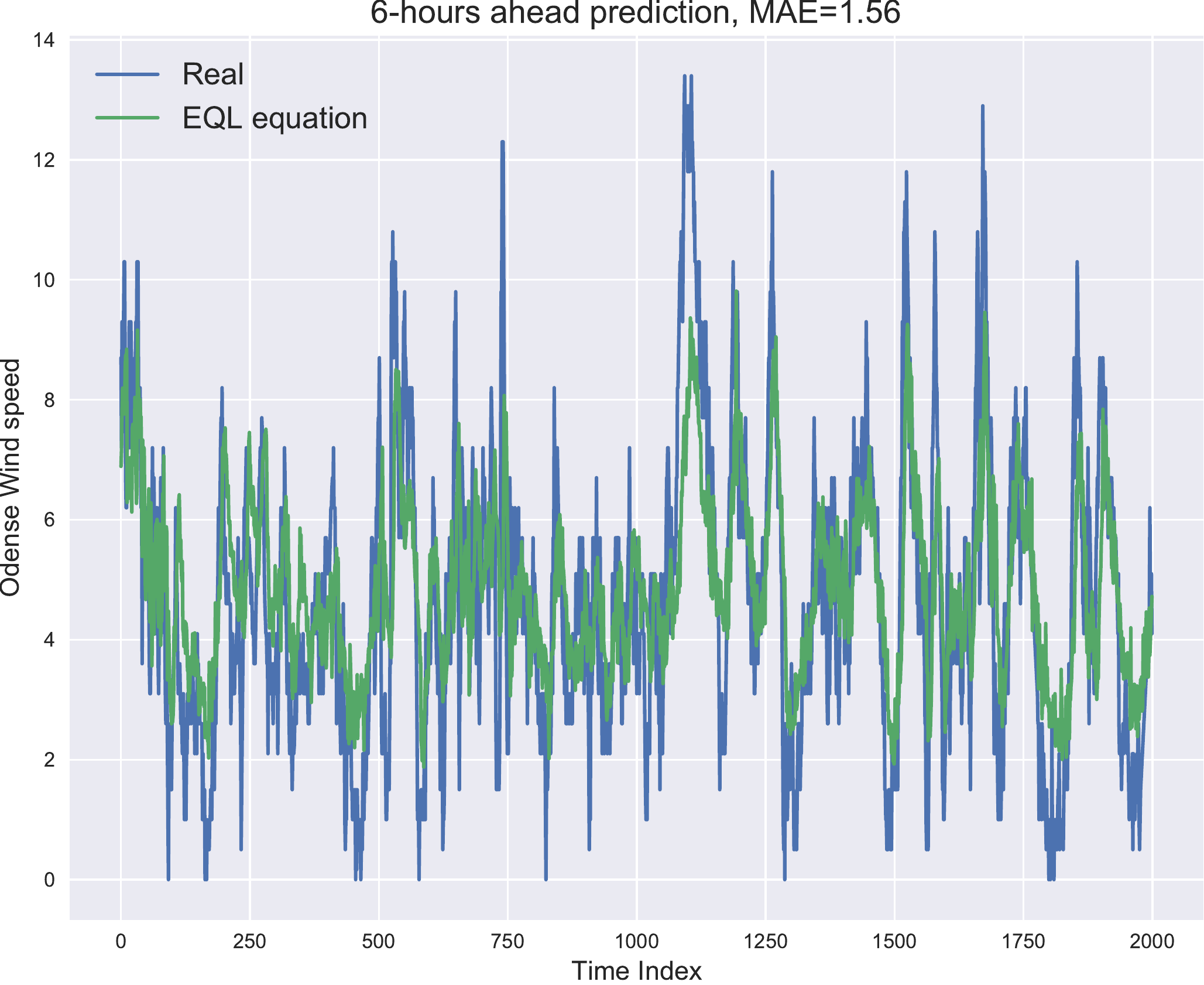}}}
\\\vskip 0.5pt plus 0.25fil
\subfloat[]{{\includegraphics[scale=0.4]{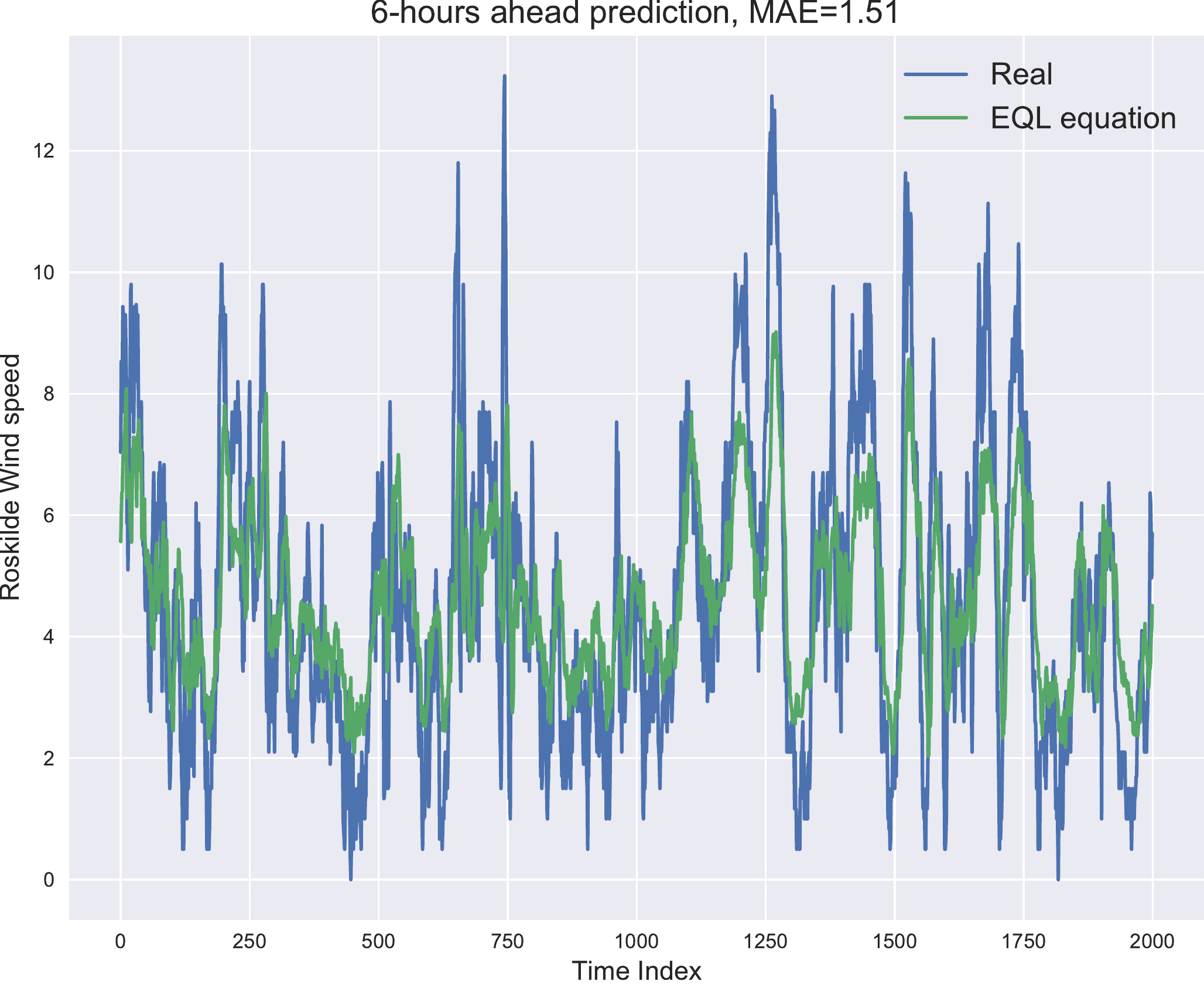}}}
\caption{Actual vs. equation prediction of the wind speed for the cities of (a) Esbjerg, (b) Odense, and (c) Roskilde.}
\label{fig:real_vs_prediction}
\end{figure}

The obtained analytical equation for each target city is tabulated in Table \ref{tab:formulas_city}. For readability purposes, the constants within the formulas have 2 decimal places, while they can have up to 16 decimal places. For the same purpose, we replace some repeated parts of the formula that contain the relevant input features by the variables $\Theta$, $\Delta$, and $\Gamma$ for the cities of Esbjerg, Odense, and Roskilde, respectively. The value of each of these variables is shown in equations \ref{eq:theta}, \ref{eq:delta}, and \ref{eq:gamma}. These equations contain a total number of 6 input features that proved to be relevant for the wind speed prediction of the 3 target cities. The input features are denoted by $x$, $y$, $z$, $\alpha$, $\beta$, and $\gamma$. An explanation of the meaning of these input features is given in Table \ref{tab:variables_explanation}. In this table, it should be noted that a higher lag value corresponds to a more recent lag in time. A plotting of the real versus the prediction from the analytical equation for each target city is show in Fig. \ref{fig:real_vs_prediction}.
\begin{table}[H]
    \centering
    \caption{Discovered analytical equation for each target city for 6h ahead wind speed predictions.}
    \label{tab:formulas_city}
    \resizebox{\columnwidth}{!}{%
    \begin{tabular}{c c}\Xhline{3\arrayrulewidth}
    \multirow{2}{*}{\textbf{City}}&
    \multirow{2}{*}{\textbf{Analytical expression}}\\
    &  \\\Xhline{3\arrayrulewidth}
       Esbjerg & \makecell{$0.32{(1-0.08sin(\Theta))}^2-0.1sin(\Theta)-0.64sin(2.12sin(\Theta)-1.37)-$\\$0.83sin(2.61sin(\Theta)+5.69)-1.36$}\\
       \makecell{\\Odense} & \makecell{\\$-2.99{(1+\frac{0.16}{\Delta})}^2+0.59-\frac{1.16}{\Delta}$}\\
       \makecell{\\Roskilde} & \makecell{\\$-14.42(-0.20-\frac{0.10}{\Gamma})*(0.05+\frac{0.61}{\Gamma})-0.02* {(1+\frac{0.67}{\Gamma})}^2-1.21$}\\\\ \hline
        \Xhline{3\arrayrulewidth}
    \end{tabular}
    }
\end{table}
\begin{equation}\label{eq:theta}
    \Theta = 0.04x-0.34y
\end{equation}
\begin{equation}\label{eq:delta}
\Delta = e^{-0.29x+0.13z+0.033\alpha+0.8y+0.79\beta+0.08\gamma}+1
\end{equation}
\begin{equation}\label{eq:gamma}
    \Gamma = e^{-0.15z-0.13\alpha-0.26y-0.44\gamma}+1
\end{equation}
\begin{table}[H]
    \centering
    \caption{Explanation of the variables used in the discovered equations.}
    \label{tab:variables_explanation}
    \begin{tabular}{c c c c}\Xhline{3\arrayrulewidth}
    \multirow{2}{*}{\textbf{Variable}}&
    \multirow{2}{*}{\textbf{City}}&\multirow{2}{*}{\textbf{Feature}}&
    \multirow{2}{*}{\textbf{Lag}}\\
    & & & \\\Xhline{3\arrayrulewidth}
       $x$ & Esbjerg & Wind speed & 3\\
       $y$ & Esbjerg & Wind speed & 4\\
       $z$ & Aalborg & Wind speed & 4\\
       $\alpha$ & Aarhus & Wind speed & 4\\
       $\beta$ & Odense & Wind speed & 4\\
       $\gamma$ & Roskilde & Wind speed & 4\\
        \hline
        \Xhline{3\arrayrulewidth}
    \end{tabular}
\end{table}

From Table \ref{tab:variables_explanation}, a first general pattern is that the most recent lag seems critical to the 6 hours ahead wind speed prediction. Indeed, most of the variables correspond to a lag of 4, except the variable $x$ that belongs to a lag of 3. Furthermore, we can observe that the 5 input cities contribute to the wind speed prediction of the target cities, but in a distinct way for each city. Moreover, only the wind speed as weather feature proved to be relevant within the analytical equations. If we combine the information given by Tables \ref{tab:formulas_city} and \ref{tab:variables_explanation}, and equations \ref{eq:theta}, \ref{eq:delta}, and \ref{eq:gamma}, we can perform an analysis on the relevant input features for each target city. If we consider the target city of Esbjerg, the relevant features concern the wind speed of the same city that belong to the two most recent lags (i.e. lags of 3 and 4). These input features are the variables $x$ and $y$. Concerning the target city of Odense, the 6  variables are relevant to the wind speed prediction. This makes sense since Odense is geographically located between Esbjerg and Roskilde, while being close to Aarhus as well. For the city of Roskilde, only the most recent wind speed of Aalborg, Aarhus, Esbjerg and Rosklide is  relevant. Another interesting observation is that while all the predictions concern the wind speed feature, the formulas for Odense and Roskilde contain the exponential function while the one for Esbjerg incorporates the sine function only. It should be added that although the exponential function was not part of the activation functions, the sigmoid function was included. Therefore the exponential function can still appear among the final analytical equation.
This indicates a possible relation between the sine and exponential functions (e.g. Euler's formula). 
If we consider Fig. \ref{fig:real_vs_prediction}, Table \ref{tab:formulas_city} and equation \ref{eq:delta}, we can observe that the most complex equation yields the worst Mean Average Error (MAE). This finding is in line with Occam's razor principle, which favors parsimonious expressions \cite{heylighen1997occam}. Similarly, the analytical equation for Roskilde contains less terms than the one for Esbjerg and yields a better MAE.

\section{Discussion} \label{sec:discussion}
\subsection{Sensitivity analysis on the sparsity}
\begin{figure}[!htbp]
\centering
\includegraphics[width=\columnwidth]{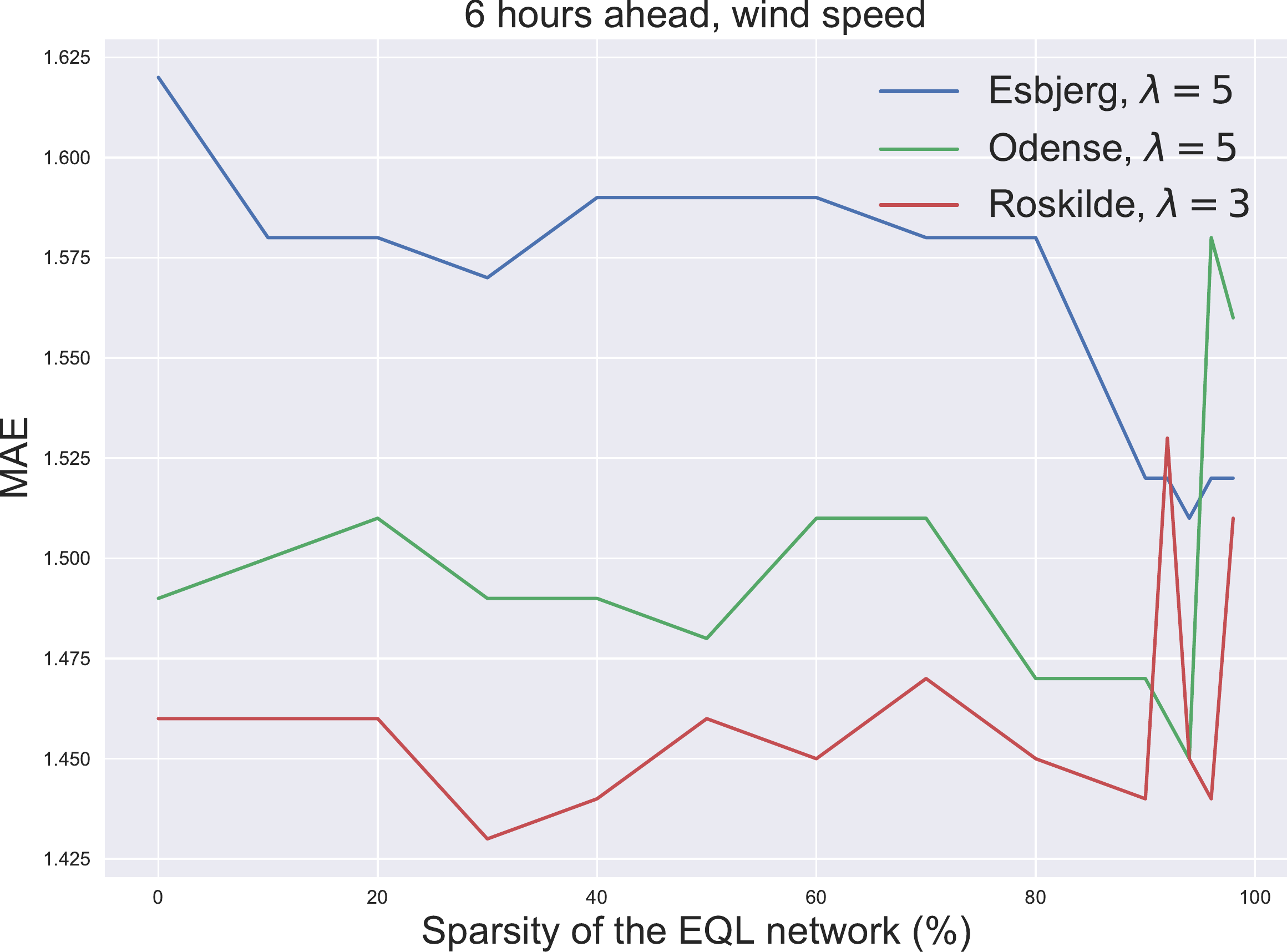}
\caption{Sparsity analysis on the cities of Esbjerg, Odense, and Roskilde.}
\label{fig:sparsity_vs_mae}
\end{figure}

Here we analyze the sensitivity of the MAE with respect to the sparsity of the network. The sparsity here is defined as the number of zero-valued weights divided by the total number of weights. For every city, we first use the pretrained model that yielded the best performance from training of the first phase. Then we keep performing only the phase two training using different thresholds to get different levels of sparsity (e.g. from 0\% up to 98\% sparsity). This analysis is shown in Fig. \ref{fig:sparsity_vs_mae}. We first observe that while Odense and Roskilde exhibit a predictable behavior, this is not the case for the city of Esbjerg. Indeed, for the cities of Odense and Roskilde, the MAE suddenly increases the more we get closer to a complete network sparsity. However, this pattern is the opposite for the city of Esbjerg, since the MAE keeps decreasing the sparser we get. This scenario depicts an ideal pattern in the context of symbolic regression, since we effectively want to get rid of unimportant weights within the network, while improving the performance. Furthermore, we observe that the optimal sparsity levels of the cities of Esbjerg, Odense, and Roskilde are 94\%, 94\%, and 30\%. We assume that Roskilde is much more affected by sparsity than the other cities because of the $\lambda$ value used during the first phase of the training. Indeed, for the cities of Esbjerg and Odense, a $\lambda$ of $5$ during regularization has been selected because it empirically yielded the best results. Similarly for Roskilde, the optimal  $\lambda$ value is $3$. Therefore, at the end of phase 1 training of Esbjerg and Odense, weights of lesser importance had lower magnitude, thus not affecting the performance after thresholding and retraining. We assume the opposite behavior happened for the city of Roskilde.

\subsection{Comparison with the baseline model}
This section compares the results obtained by the EQL network with the those of \cite{mehrkanoon2019deep2}. As stated previously, the successful model that has been used on the same dataset is based on a 3D convolutional network. A first difference is based on the explanability of the approach. While the work in \cite{mehrkanoon2019deep2} applies traditional deep learning to get a good extrapolation ability, the trained model is a black box model and does not emphasize the important input features. On the other hand, the EQL approach yields a mathematical expression that uncovers the important input features. Indeed, the trained 3d-CNN uses all 80 input features to make inference while the EQL model needs only a maximum of 7.5\% of the input features (e.g. 6 of them). Therefore the EQL approach is much more desirable in terms of explanability. There is also a drastic change of the inference time. Table \ref{tab:perf_comp} tabulates the inference time in seconds as well as the performance of both models for the three studied target cities. It should be noted that this inference time is obtained over 5000 samples, and averaged over 10 trials to get statistically significant results. A first observation is that the EQL-based models take much less time to make inference than the 3d-CNN model. There is approximately a 43$\times$, 29$\times$, and 28$\times$ speedup for the cities of Esbjerg, Odense, and Roskilde, respectively. Esbjerg is much faster to infer thanks to the low number of terms used inside the $\sin(\cdot)$ function. One can notice that the EQL inference time of each city is different since there is a distinct analytical equation for each target city. However, it takes about the same time to infer any city using the 3d-CNN model, since the number of parameters is constant.

Finally, a crucial comparison concerns the performance of the two approaches. Based on the training procedure of the EQL network and the sparsity constraint to yield compact analytical equations, a lower performance is expected from the EQL network. If we consider Table \ref{tab:perf_comp}, it should be noted that the MAE in the third column shows the best MAE across all possible levels of sparsity. Concerning the MAE shown in the  4\textsuperscript{th} column, called MAE$^*$, this corresponds to a sparsity level of 98\% (e.g. the one that yields the formulas shown in Table \ref{tab:formulas_city}). If we focus on the best MAE for both models, the city of Roskilde benefits from the EQL approach, with a decrease of the MAE from 1.48 to 1.43. However, the EQL approach is detrimental to the cities of Esbjerg and Odense, with a large MAE increase that concerns the latter city. Overall, even if the EQL approach might result is a slight decrease of performance, it is very beneficial in terms of explainability and inference.

\begin{table}[H]
    \centering
    \caption{Performance and inference time comparison with the baseline model for the 3 target cities.}
    \label{tab:perf_comp}
    \resizebox{0.7\columnwidth}{!}{%
    \begin{tabular}{l l l l l}\Xhline{3\arrayrulewidth}
    \multirow{2}{*}{\textbf{Model}}&
    \multirow{2}{*}{\textbf{City}}&\multirow{2}{*}{\textbf{MAE}}
    &\multirow{2}{*}{\textbf{MAE$^*$}}&\multirow{2}{*}{\textbf{Inference time (s)}}\\
    & & & &\\\Xhline{3\arrayrulewidth}
       EQL & Esbjerg & 1.51 & 1.52 &0.0879\\
        & Odense & 1.45 & 1.56 & 0.129\\
        & Roskilde & 1.43 & 1.51& 0.132\\\hline
        3d-CNN & Esbjerg & 1.40 & -&3.73\\
        & Odense & 0.62 & -&3.77\\
        & Roskilde & 1.48 & -&3.74\\
        \hline
        \Xhline{3\arrayrulewidth}
    \end{tabular}
    }
\end{table}

\subsection{Analysis of the 2\textsuperscript{nd} training phase}
\begin{figure}[!htbp]
\centering
\subfloat[]{{\includegraphics[scale=0.4]{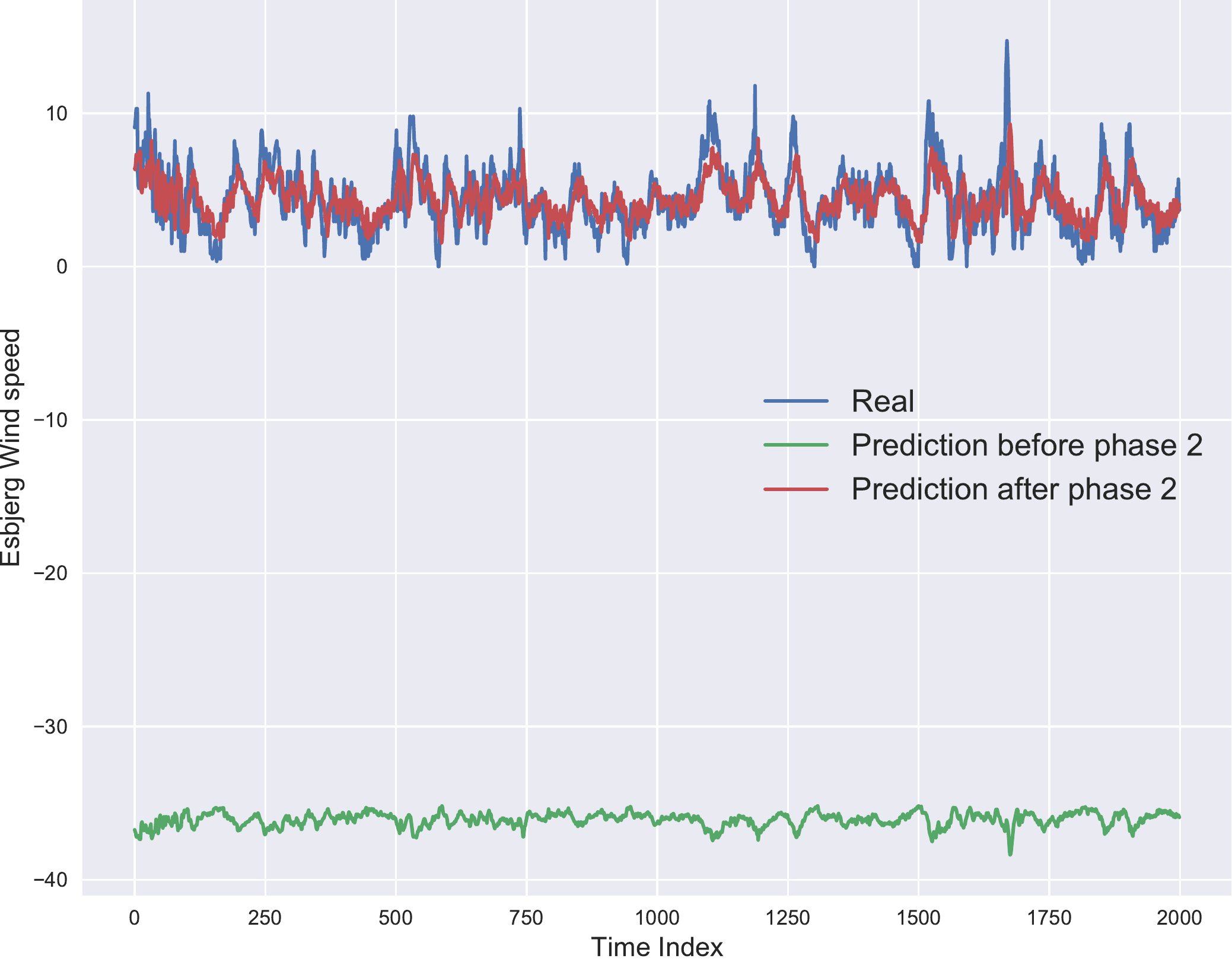}}}
\\\vskip 0.5pt plus 0.25fil
\subfloat[]{{\includegraphics[scale=0.4]{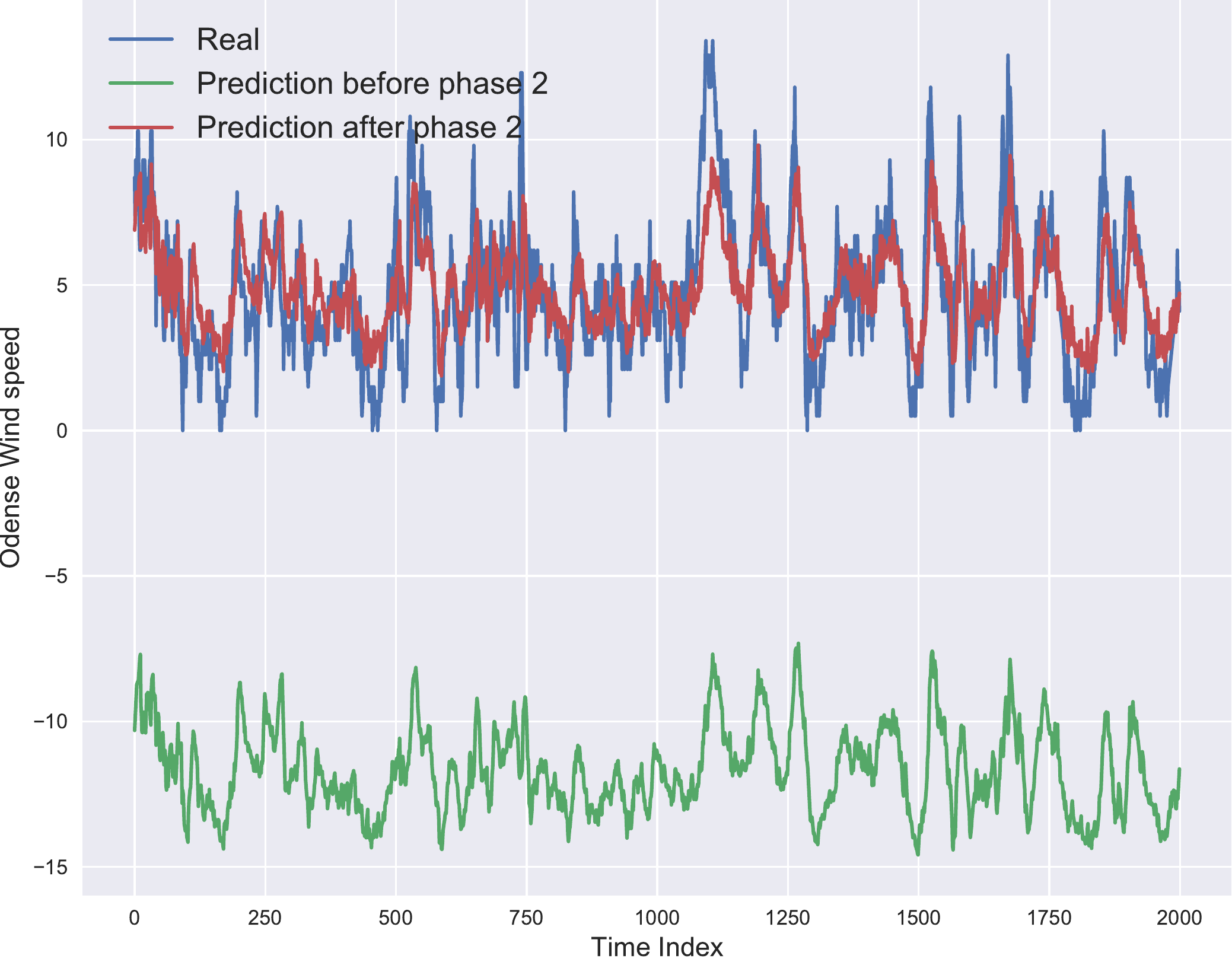}}}
\\\vskip 0.5pt plus 0.25fil
\subfloat[]{{\includegraphics[scale=0.4]{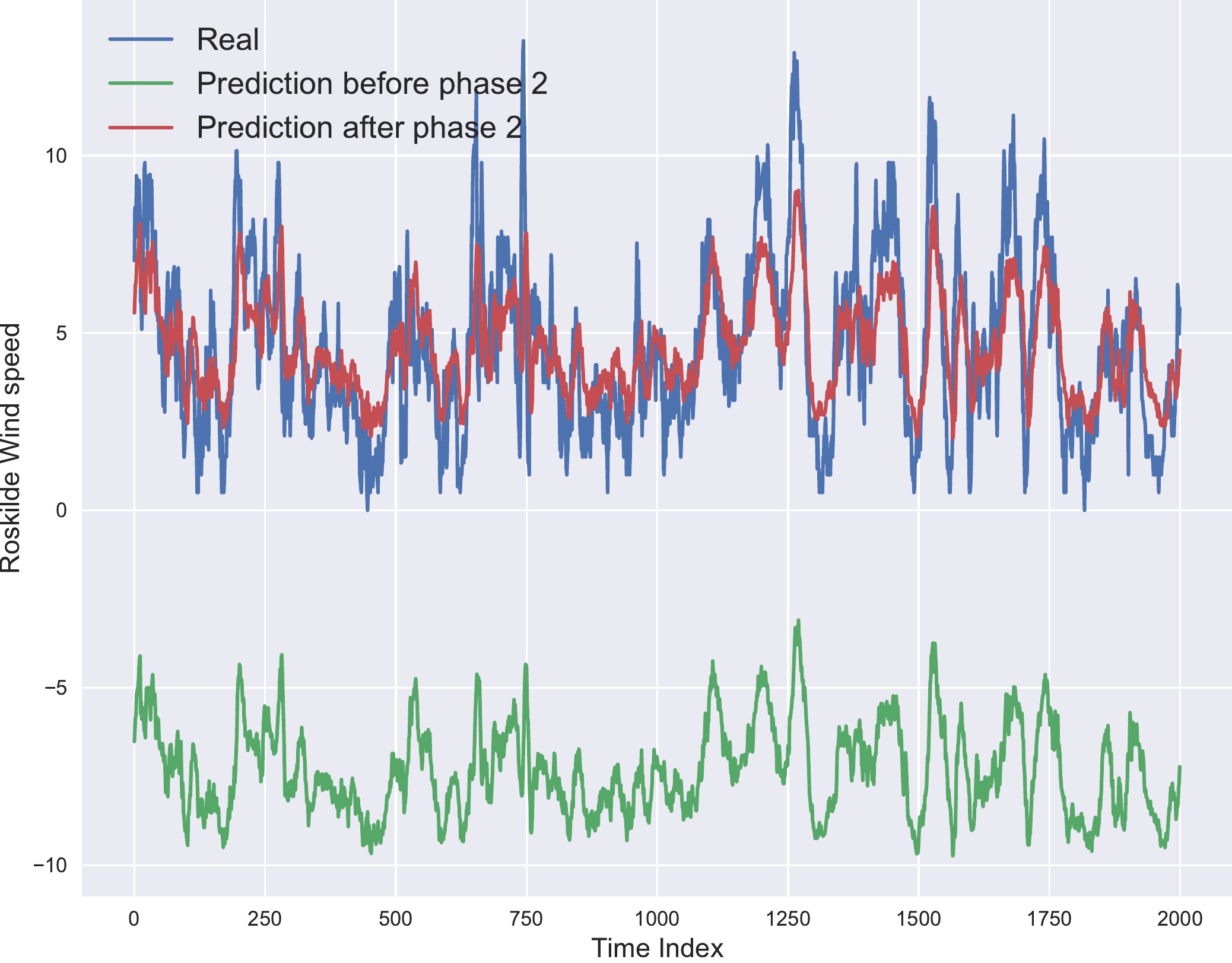}}}
\caption{Real vs. prediction at different phases of the training after thresholding  for the cities of (a) Esbjerg, (b) Odense, and (c) Roskilde.}
\label{fig:phase2_analysis}
\end{figure}

\begin{figure}[!htbp]
\includegraphics[width=\columnwidth]{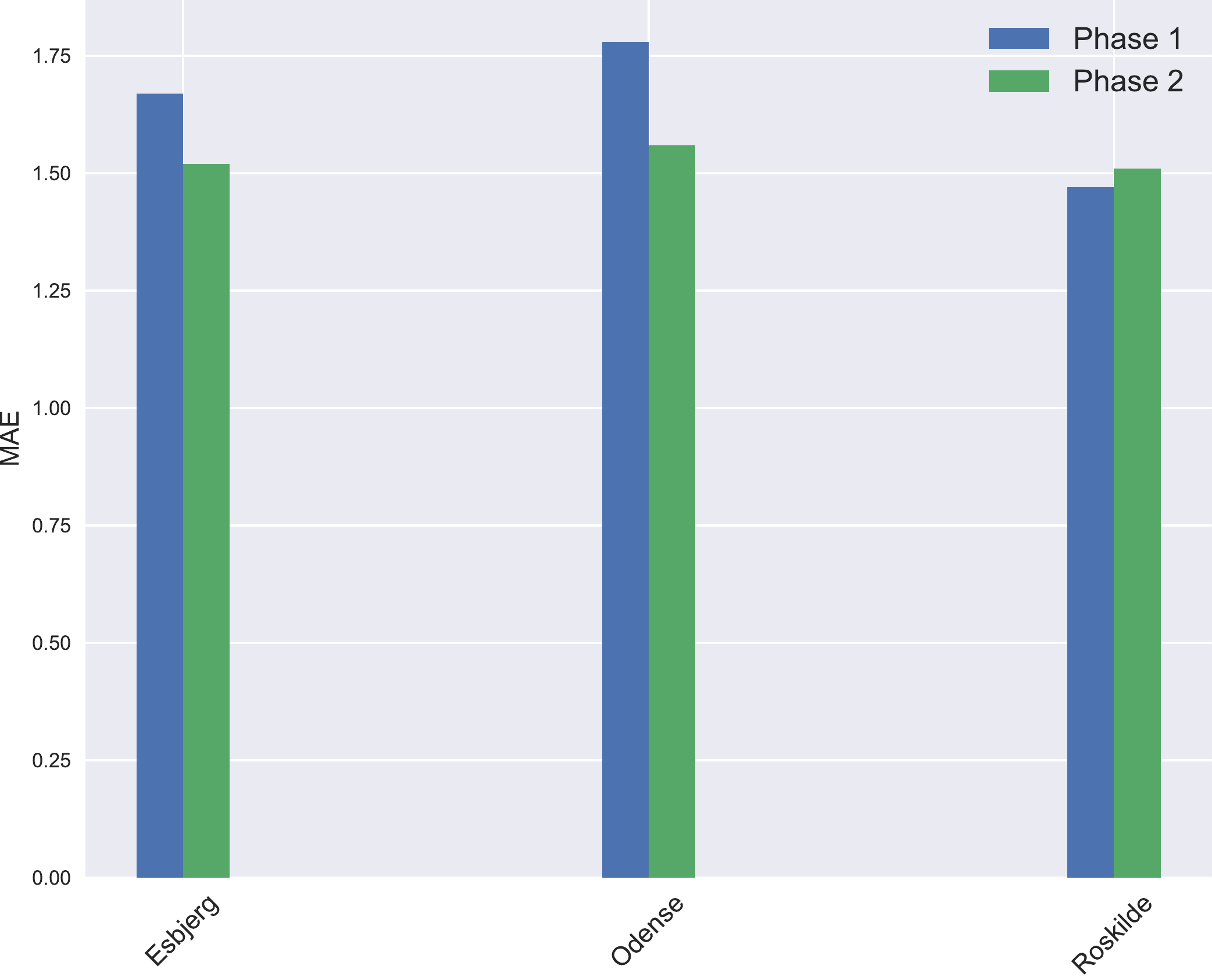}
\caption{Comparison of the MAE between the 2 training phases for the  cities of Esbjerg, Odense, and Roskilde.}
\label{fig:p1_p2_hist}
\end{figure}

As stated in section \ref{ssec:eql_layer}, the 2\textsuperscript{nd} training phase consists of performing a thresholding that will decide on the sparsity of the network. A regular training is then conducted without regularization while freezing the zero-valued weights. Fig. \ref{fig:phase2_analysis} shows the real data for each city, as well as the prediction before and after the phase 2 training. We should note that a thresholding that results in a 98\% network sparsity is performed. Interestingly enough, we can observe from the training that concerns Odense and Roskilde that the  2\textsuperscript{nd} training phase effectively learns an affine transformation (e.g. translation transformation). Concerning the city of Esbjerg, it seems that this training learns a more complex affine transformation consisting of a combination of translation and scaling. We assume here that setting a value of 0 to the lower magnitude weights induces an affine transformation between the prediction data and the real data. This observation is in-line with the observations made in \cite{champion2019data}. Fig. \ref{fig:p1_p2_hist} shows a comparison of the descaled MAE after the 2 training phases for the 3 target cities. While the 2\textsuperscript{nd} phase is beneficial for both the cities of Esbjerg and Odense since it decreases the MAE, it is not the case for Roskilde. For this particular city, there is a slight MAE increase.

\section{Conclusion}\label{sec:conclusion}
In this paper, an EQL based architecture is investigated to perform wind speed forecasting. We show that the EQL-based approach can yield a mathematical expression that results in an accurate enough wind speed prediction. The major added-value of this approach is its explanability since the compact mathematical expression incorporates the relevant input features as well as their relations. Another advantage of this approach is the inference time speedup compared to traditional neural networks. It has also been shown that $L_{0.5}^*$ regularization is beneficial in the context of wind speed prediction.
Finally, an interesting observation about the role of the second training phase, since it has been shown that this training phase is effectively learning an affine  transformation. This work has shown that an EQL-based model is feasible for analyzing high-dimensional chaotic data such as weather observation. Both the code and data can be found at the following links, respectively: \href{https://www.github.com/IsmailAlaouiAbdellaoui/EQL-Wind-Speed-Forecasting/}{github.com/IsmailAlaouiAbdellaoui/EQL-Wind-Speed-Forecasting/} and \href{https://www.sites.google.com/view/siamak-mehrkanoon/code-data?authuser=0}{sites.google.com/view/siamak-mehrkanoon/code-data?authuser=0}.

\section*{Acknowledgment}
Simulations were performed with computing resources granted by RWTH Aachen University and Cloud TPUs from Google's TensorFlow Research Cloud (TFRC).

\bibliography{Main}

\end{document}